\documentclass{article}

\usepackage[sglblindworkshop, final]{neurips_2026}

\usepackage[utf8]{inputenc}
\usepackage[T1]{fontenc}
\usepackage{hyperref}
\usepackage{url}
\usepackage{booktabs}
\usepackage{amsmath,amssymb,amsfonts}
\usepackage{nicefrac}
\usepackage{microtype}
\usepackage{xcolor}
\usepackage{graphicx}
\usepackage{multirow}
\usepackage{float}
\hypersetup{hidelinks}

\workshoptitle{Second Workshop on ML$\times$OR: Mathematical Foundations and Operational Integration of Machine Learning for Uncertainty-Aware Decision-Making}

\title{Scenario Augmentation for Rare Financial Events via Multimodal Latent Diffusion}

\author{
  Mahdi Arab Loodaricheh$^{1}$ \qquad
  Anita Raja$^{2}$ \qquad
  Indranil SenGupta$^{2}$ \\
  {\small
  $^{1}$The Graduate Center, CUNY
  \qquad
  $^{2}$Hunter College, CUNY
  } \\
  {{\small Corresponding author:
\texttt{mahdi.arabloodaricheh41@cuny.edu}}}
}
\begin{document}

\maketitle

\begin{abstract}
Extreme market movements are scarce, heterogeneous, and difficult to learn from imbalanced historical data. We study whether generative augmentation can improve the representation and
classification of CVaR-labeled rare financial windows while retaining the
complementary structure exposed by different financial representations. Each market window is expressed through three aligned modalities: structured market variables, a spatial image transformation, and a sequence of portfolio returns. Separate variational autoencoders map the modalities to compact latent codes, and a class-conditional diffusion model learns their joint latent distribution. Conditional Value-at-Risk (CVaR) provides a risk-aware definition of rare windows and directs generation toward the minority regime. Across four window lengths on U.S. technology equities, the resulting synthetic samples improve fidelity and support relative to tabular diffusion baselines. An XGBoost classifier trained with real and generated samples attains AUCs of 0.953--0.960, exceeding both tabular generators and the corresponding real-only baselines. These results show that multimodal latent generation can provide useful synthetic stress scenarios for data-scarce financial risk models.
\end{abstract}

\section{Introduction}

Data-driven financial risk models face a structural scarcity problem: the market regimes of greatest operational importance are precisely those observed least often. Extreme losses are infrequent, high-dimensional, and heterogeneous, while standard predictors are dominated by normal-market observations~\citep{christoffersen1998,basu2022novel}. This imbalance is consequential for uncertainty-aware decision support because tail-risk estimates and warning systems depend on a faithful representation of stressed conditions rather than average behavior alone.

Generative augmentation offers a natural response, but financial windows contain several kinds of structure. Raw tabular variables preserve feature-level values; spatial transformations expose local cross-feature and cross-asset relations; and return sequences retain temporal dynamics. Multimodal learning can combine such complementary representations~\citep{Baltrusaitis2019}, while diffusion models provide stable distributional coverage for complex data~\citep{Ho2020}. Existing tabular diffusion methods such as TabDDPM~\citep{kotelnikov2023tabddpm} and TabSyn~\citep{zhang2024mixedtype} are strong generators, but they model only the structured view. MAGE-ID similarly uses multimodal latent diffusion for imbalanced-data augmentation~\citep{ArabLoodaricheh2026MAGEID}; we extend this approach to financial sequences and CVaR-defined rare events.

We therefore investigate a risk-aware multimodal augmentation framework for rare financial events. A market window is encoded as synchronized tabular, image, and sequence modalities. Modality-specific variational autoencoders (VAEs)~\citep{Kingma2014} produce compact representations that are concatenated and modeled by class-conditional latent diffusion. Rare-event labels are defined using CVaR, connecting the conditioning mechanism to a coherent tail-risk measure~\citep{Rockafellar2000}. The generated minority samples are evaluated both distributionally and through their utility for a held-out rare-event classifier.

Our contribution is threefold: (i) a multimodal latent representation of aligned structural, spatial, and temporal financial information; (ii) class-conditional diffusion for targeted synthesis of CVaR-defined rare events; and (iii) an evaluation showing that the generated samples improve synthetic-to-real transfer and real-data augmentation over two tabular diffusion baselines. The study focuses on predictive decision support; it does not claim to evaluate an end-to-end portfolio policy.

\section{Risk-aware multimodal generation}
\label{sec:method}

\begin{figure}[t]
  \centering
  \includegraphics[width=0.92\linewidth]{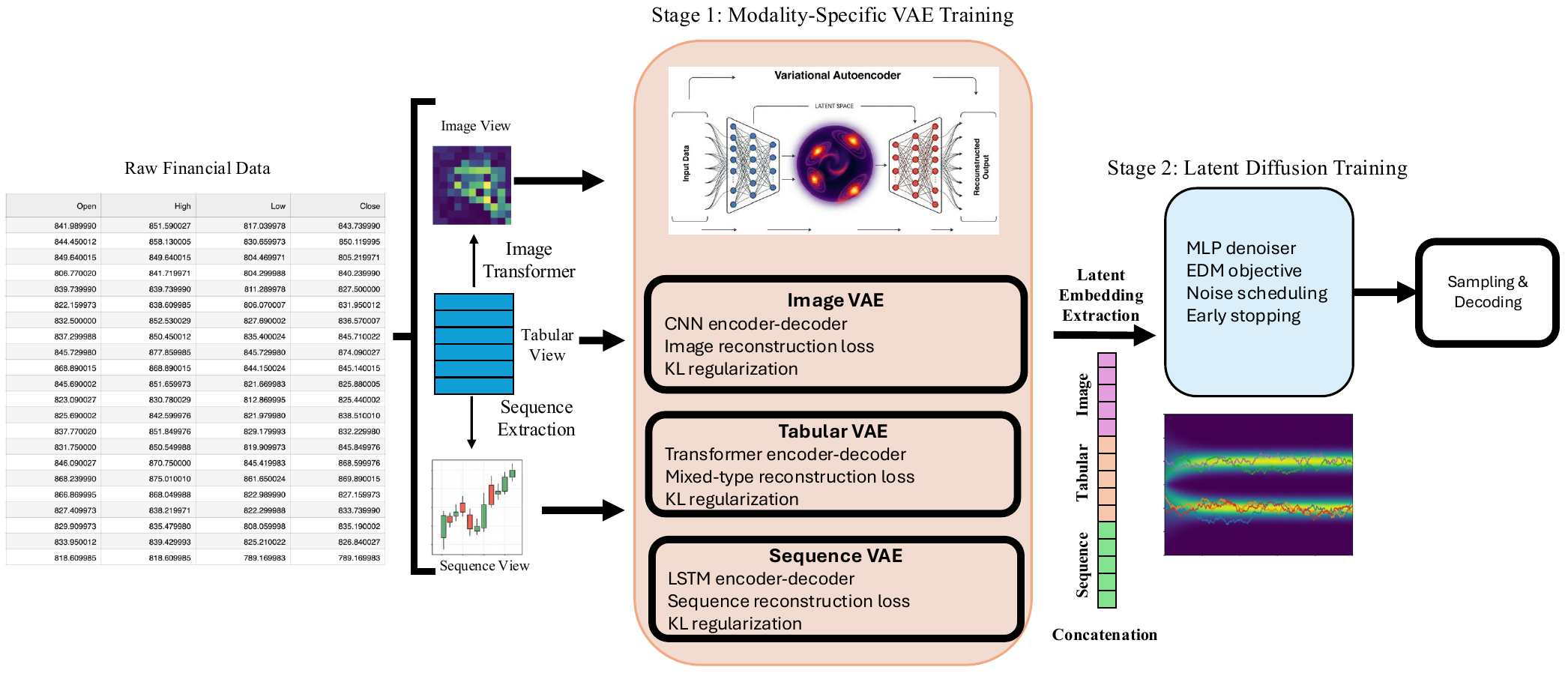}
  \caption{Aligned tabular, image, and sequence views are encoded by modality-specific VAEs, modeled by class-conditional latent diffusion, and decoded into synthetic windows.}
  \label{fig:framework}
\end{figure}

\paragraph{Financial windows and risk labels.}
We construct windows of length $N\in\{5,6,7,8\}$ from daily observations of eight major U.S. technology equities. For every equity, the retained variables are open price, close price, and volume, and the overnight gap is defined as $\mathrm{Gap}^{(i)}_t=\mathrm{Open}^{(i)}_t-\mathrm{Close}^{(i)}_{t-1}$.
Features are min--max normalized using training-fold statistics, and equities are ordered by CAPM beta. For equal-weight return $R_t$, define $L_t=-R_t$ and \mbox{$\mathrm{CVaR}_{\alpha}=\mathbb{E}[L_t\mid L_t>\mathrm{VaR}_{\alpha}]$}.
A window is labeled rare ($y=1$) when its CVaR exceeds the prescribed risk
threshold and normal ($y=0$) otherwise, following the risk-aware construction
of~\citet{Lubdhak}. Risk awareness enters through the CVaR-derived label; the VAE encoders
themselves remain risk-agnostic. The label characterizes the same $N$-day window represented
by the input modalities rather than a subsequent prediction horizon. Therefore,
the downstream task is contemporaneous rare-window classification, not
future-event forecasting. Depending on $N$, rare windows comprise
9.09--13.65\% of the data.We focus on five- to eight-day windows because rare market movements often
develop over compact multi-day periods. These horizons provide sufficient
temporal context to represent short-term return dynamics while avoiding longer
windows that may dilute local rare-event signals and reduce the number of
complete training samples.

\paragraph{Aligned modalities.}
The \emph{tabular modality} concatenates normalized close, volume, and gap
variables across assets and days. For eight equities, this produces
\(F=3|\mathcal{K}|N\) features, ranging from 120 features for \(N=5\) to 192
features for \(N=8\). The \emph{image modality} applies
DeepInsight~\citep{sharma2019deepinsight}. The feature layout is learned using
only the training fold by embedding features in two dimensions according to
their similarity across training samples. The resulting coordinates are mapped to a fixed $12\times12$ grid; when multiple
features occupy the same pixel, their values are averaged. For $N=7$ and $N=8$,
the 168 and 192 features exceed the 144-pixel grid capacity, making collisions
and averaging unavoidable; consequently, the image is treated as a lossy auxiliary
view, while the tabular branch retains the complete feature-level representation.
The \emph{sequence modality} contains the
standardized equal-weight portfolio-return trajectory over the same window.
Thus, the three synchronized modalities expose feature-level, spatial, and
temporal structure without introducing unobserved market information.

\paragraph{Modality-specific representation learning.}
A Transformer VAE models the tabular view, a convolutional VAE models the image, and a bidirectional LSTM VAE models the return sequence. For modality $m\in\{\mathrm{tab},\mathrm{img},\mathrm{seq}\}$, the encoder produces $q_{\phi_m}(z_m\mid x_m)$ and the decoder reconstructs $\hat{x}_m$. Stage 1 minimizes reconstruction error and KL regularization, together with pairwise alignment of synchronized latent codes:
\begin{align}
\mathcal{L}_{\mathrm{VAE}}={}&\sum_m\left[\lVert x_m-\hat{x}_m\rVert_2^2+
\beta D_{\mathrm{KL}}\!\left(q_{\phi_m}(z_m\mid x_m)\,\|\,\mathcal{N}(0,I)\right)\right] \nonumber\\
&+\lambda_{\mathrm{align}}\left(\lVert z_{\mathrm{tab}}-z_{\mathrm{img}}\rVert_2^2+
\lVert z_{\mathrm{tab}}-z_{\mathrm{seq}}\rVert_2^2\right).
\end{align}
Equal latent dimensionalities give the three modalities comparable
representational capacity before fusion. The image and sequence latents are
standardized using training-fold means and standard deviations. The tabular
latent is additionally partially whitened to reduce strong correlations and
anisotropy while preserving useful structure. These transformations prevent
differences in latent scale or covariance from causing one modality to dominate
the concatenated representation and provide a better-conditioned space for
diffusion training. During generation, the inverse transformations are applied
before the modality-specific decoders reconstruct the synthetic observations.
The normalized codes are concatenated as
\(z=[z_{\mathrm{tab}},z_{\mathrm{img}},z_{\mathrm{seq}}]\).

\paragraph{Conditional latent diffusion.}
With the VAE parameters fixed, an MLP denoiser is trained on noisy versions of $z$ using a continuous EDM-style schedule. A label embedding conditions the denoiser on normal or rare-event generation. Gaussian noise serves only as the forward corruption process; tail-risk information enters the reverse process through the CVaR-derived class label. Writing $z_t$ for a noisy latent and $\epsilon$ for Gaussian noise, Stage 2 minimizes $\mathcal{L}_{\mathrm{diff}}=\mathbb{E}_{z,\epsilon,t}\!\left[\lVert\epsilon_\theta(z_t,t,y)-\epsilon\rVert_2^2\right]$.
Sampling begins from Gaussian noise, applies the reverse process, separates the generated latent blocks, reverses normalization, and decodes each modality. The generated tabular component is used for downstream rare-event augmentation; the synchronized image and sequence components constrain learning of the joint latent distribution. Figure~\ref{fig:framework} summarizes the pipeline.

\section{Experimental evaluation}
\label{sec:experiments}

\paragraph{Protocol.}
We evaluate $N\in\{5,6,7,8\}$ with five-fold cross-validation. Within every fold, normalization, DeepInsight placement, latent transformations, VAE training, diffusion training, and synthesis use only the training partition. The VAE and diffusion stages are each trained for 2000 epochs. We compare against the official implementations of TabDDPM and TabSyn. Synthetic fidelity and diversity are measured with Precision, Recall, Density, and Coverage (PRDC)~\citep{naeem2020reliable}.
Precision measures the proportion of generated samples supported by the
real-data manifold, while Recall measures how broadly the generated
distribution covers that manifold. Density evaluates the concentration of
synthetic samples around real observations, and Coverage measures the
proportion of real samples represented by nearby generated samples. Reporting
all four metrics is important because broad distributional coverage alone does
not guarantee that the generated observations are locally realistic. Downstream classification utility is evaluated with XGBoost under three protocols: \emph{Synth$\rightarrow$Real}, trained only on generated samples and tested on held-out real data; \emph{Real+Synth$\rightarrow$Real}, trained on augmented real data and tested on held-out real data; and \emph{Real$\rightarrow$Real}, the unaugmented reference. All reported values are fold means and standard deviations from the existing evaluation.

\begin{table}[t]
\centering
\caption{Existing five-fold results (mean $\pm$ standard deviation). Bold marks the best generated-data value for each $N$.}
\label{tab:results}
\fontsize{7}{7.8}\selectfont
\textbf{(a) Generative quality (PRDC).}\\[-1pt]
\setlength{\tabcolsep}{2.8pt}
\begin{tabular}{@{}clcccc@{}}
\toprule
$N$ & Generator & Precision & Recall & Density & Coverage \\
\midrule
\multirow{3}{*}{5}
& TabDDPM & $0.3972\pm0.0376$ & $0.7432\pm0.0152$ & $\mathbf{0.3154\pm0.0400}$ & $0.3792\pm0.0317$ \\
& TabSyn & $0.2133\pm0.0375$ & $0.8018\pm0.0129$ & $0.1189\pm0.0297$ & $0.1821\pm0.0288$ \\
& \textbf{Ours} & $\mathbf{0.4058\pm0.0512}$ & $\mathbf{0.8640\pm0.0084}$ & $0.2373\pm0.0436$ & $\mathbf{0.4156\pm0.0340}$ \\
\midrule
\multirow{3}{*}{6}
& TabDDPM & $0.2309\pm0.0589$ & $0.7548\pm0.0145$ & $\mathbf{0.1679\pm0.0469}$ & $0.2490\pm0.0369$ \\
& TabSyn & $0.1024\pm0.0341$ & $0.8566\pm0.0544$ & $0.0486\pm0.0201$ & $0.0853\pm0.0323$ \\
& \textbf{Ours} & $\mathbf{0.2708\pm0.0431}$ & $\mathbf{0.8898\pm0.0181}$ & $0.1289\pm0.0556$ & $\mathbf{0.2786\pm0.0387}$ \\
\midrule
\multirow{3}{*}{7}
& TabDDPM & $0.1161\pm0.0581$ & $0.7482\pm0.0266$ & $0.0816\pm0.0410$ & $0.1396\pm0.0579$ \\
& TabSyn & $0.0065\pm0.0045$ & $\mathbf{0.9887\pm0.0050}$ & $0.0024\pm0.0017$ & $0.0046\pm0.0028$ \\
& \textbf{Ours} & $\mathbf{0.3892\pm0.0366}$ & $0.5545\pm0.0233$ & $\mathbf{0.3452\pm0.0513}$ & $\mathbf{0.3327\pm0.0209}$ \\
\midrule
\multirow{3}{*}{8}
& TabDDPM & $0.0864\pm0.0327$ & $0.7605\pm0.0145$ & $0.0648\pm0.0257$ & $0.1303\pm0.0402$ \\
& TabSyn & $0.0034\pm0.0020$ & $\mathbf{0.9868\pm0.0114}$ & $0.0011\pm0.0007$ & $0.0022\pm0.0016$ \\
& \textbf{Ours} & $\mathbf{0.3301\pm0.0377}$ & $0.5634\pm0.0437$ & $\mathbf{0.2802\pm0.1687}$ & $\mathbf{0.3070\pm0.0572}$ \\
\bottomrule
\end{tabular}

\vspace{3pt}
\textbf{(b) Downstream AUC. S$\rightarrow$R is synthetic-to-real; R+S$\rightarrow$R is augmented training.}\\[-1pt]
\setlength{\tabcolsep}{3.4pt}
\begin{tabular}{@{}clccc@{}}
\toprule
$N$ & Generator & S$\rightarrow$R & R+S$\rightarrow$R & Real$\rightarrow$Real \\
\midrule
\multirow{3}{*}{5}
 & TabDDPM & $0.9057\pm0.0143$ & $0.9349\pm0.0083$ & \multirow{3}{*}{$0.9290\pm0.0112$} \\
 & TabSyn  & $0.9020\pm0.0171$ & $0.9383\pm0.0094$ & \\
 & \textbf{Ours} & $\mathbf{0.9263\pm0.0053}$ & $\mathbf{0.9550\pm0.0008}$ & \\
\midrule
\multirow{3}{*}{6}
 & TabDDPM & $0.9129\pm0.0100$ & $0.9451\pm0.0065$ & \multirow{3}{*}{$0.9342\pm0.0082$} \\
 & TabSyn  & $0.8678\pm0.0674$ & $0.9430\pm0.0044$ & \\
 & \textbf{Ours} & $\mathbf{0.9300\pm0.0148}$ & $\mathbf{0.9596\pm0.0063}$ & \\
\midrule
\multirow{3}{*}{7}
 & TabDDPM & $0.9101\pm0.0126$ & $0.9445\pm0.0085$ & \multirow{3}{*}{$0.9323\pm0.0097$} \\
 & TabSyn  & $0.8603\pm0.0737$ & $0.9425\pm0.0063$ & \\
 & \textbf{Ours} & $\mathbf{0.9151\pm0.0057}$ & $\mathbf{0.9532\pm0.0034}$ & \\
\midrule
\multirow{3}{*}{8}
 & TabDDPM & $0.8913\pm0.0134$ & $0.9372\pm0.0064$ & \multirow{3}{*}{$0.9236\pm0.0042$} \\
 & TabSyn  & $0.8151\pm0.1001$ & $0.9374\pm0.0081$ & \\
 & \textbf{Ours} & $\mathbf{0.9048\pm0.0051}$ & $\mathbf{0.9558\pm0.0058}$ & \\
\bottomrule
\end{tabular}
\end{table}




\section{Results and discussion}

\paragraph{Generative quality.}
Table~\ref{tab:results}(a) shows that the multimodal generator provides the
strongest overall balance between fidelity and distributional support. For
\(N=5\), it achieves the highest Precision, Recall, and Coverage, while
TabDDPM obtains the highest Density. A similar pattern appears for \(N=6\):
our method produces the highest Precision, Recall, and Coverage, with a
moderate reduction in Density relative to TabDDPM. These results indicate that
the generated samples cover a broader portion of the rare-event distribution
without concentrating only around a small set of observed examples. The difference becomes more pronounced for \(N=7\) and \(N=8\). TabSyn obtains
Recall values close to one, but its Precision, Density, and Coverage are nearly
zero. This combination suggests that its samples are widely dispersed and
frequently fall outside regions strongly supported by the real data. In
contrast, multimodal latent diffusion maintains substantially higher
Precision, Density, and Coverage at both window sizes. Although its Recall is
lower than that of TabSyn, it provides a more balanced trade-off between
diversity and local realism. This robustness is consistent with the use of
modality-specific encoders and latent normalization, which reduce scale and
covariance mismatch before diffusion.

\paragraph{Downstream classification utility.}
Table~\ref{tab:results}(b) shows that the proposed generator achieves the
highest synthetic-to-real AUC for all four window sizes. This indicates that
the generated observations preserve class-discriminative structure that
transfers to held-out real data, rather than only reproducing marginal feature
statistics. The benefit is stronger when synthetic samples augment the real training set.
Real-plus-synthetic AUC ranges from \(0.9532\) to \(0.9596\) and exceeds the
corresponding real-only result for every window size. For example, at \(N=6\),
augmentation increases AUC from \(0.9342\pm0.0082\) to
\(0.9596\pm0.0063\), while at \(N=8\) it increases AUC from
\(0.9236\pm0.0042\) to \(0.9558\pm0.0058\). The consistent improvements
across all horizons show that the generated rare-event samples provide useful
complementary training information under class imbalance.
Additional class counts, qualitative examples, and runtime measurements are
provided in Appendix~\ref{app:supplement}.

\section{Conclusion}

We presented a tail-risk-aware augmentation framework that models aligned
tabular, image, and sequence representations through modality-specific VAEs
and class-conditional latent diffusion. CVaR provides a risk-aware definition
of rare events, while targeted generation addresses their scarcity. Across
four window sizes, the generated samples improve synthetic-to-real transfer
and real-data augmentation relative to TabDDPM and TabSyn, demonstrating the
value of multimodal generation for uncertainty-aware financial risk modeling.
The current evaluation focuses on U.S. technology equities; future work will
consider broader markets and further analyze the contribution of each modality.

\bibliographystyle{plainnat}
\bibliography{references}

\appendix
\section{Supplementary material}
\label{app:supplement}

\subsection{Class distribution}
Table~\ref{tab:class_distribution} summarizes the class composition for each window size. Rare events account for only 9.09\%--13.65\% of the observations, confirming substantial class imbalance across all horizons and motivating targeted minority-class augmentation. The total sample count decreases slightly as $N$ increases because longer rolling windows yield fewer complete observations, while the rare-event proportion rises as longer windows capture more periods of elevated cumulative loss.
\begin{table}[h]
\centering
\caption{Class distribution across window sizes.}
\label{tab:class_distribution}
\small
\begin{tabular}{@{}cccc@{}}
\toprule
$N$ & Normal samples & Rare-event samples & Rare-event ratio \\
\midrule
5 & 5272 & 527 & 9.09\% \\
6 & 5174 & 624 & 10.76\% \\
7 & 5085 & 712 & 12.28\% \\
8 & 5005 & 791 & 13.65\% \\
\bottomrule
\end{tabular}
\end{table}

\subsection{Representative transformed and generated images}
\begin{figure}[H]
  \centering
  \includegraphics[width=0.88\linewidth]{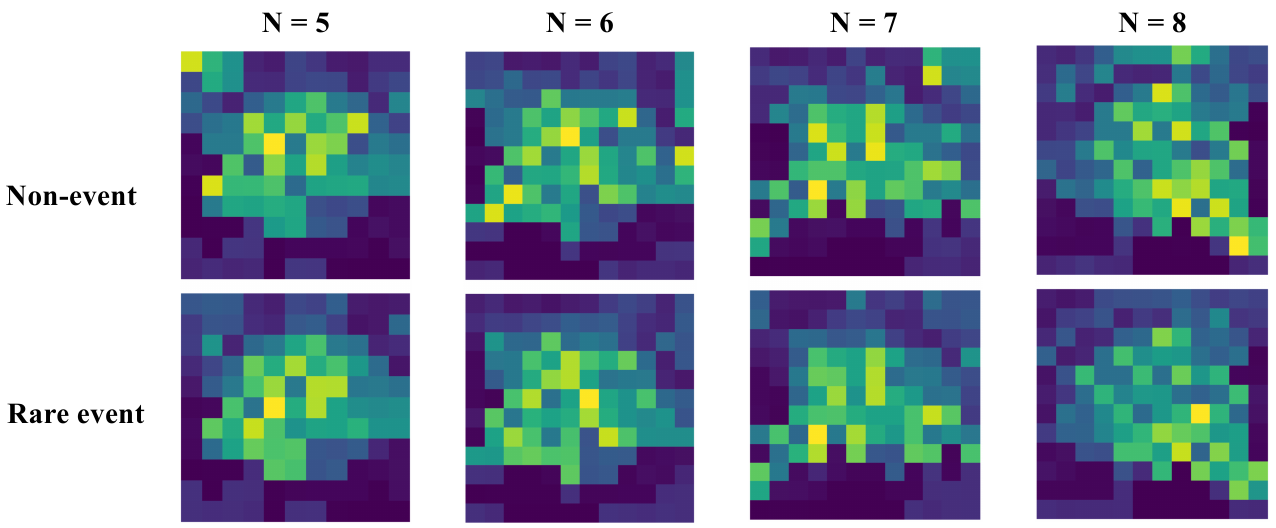}
  \caption{Representative DeepInsight images for normal and rare-event windows across $N\in\{5,6,7,8\}$. Pixels encode normalized close, volume, and gap variables arranged by feature similarity.}
  \label{fig:deepinsight}
\end{figure}

\begin{figure}[H]
  \centering
  \includegraphics[width=0.70\linewidth]{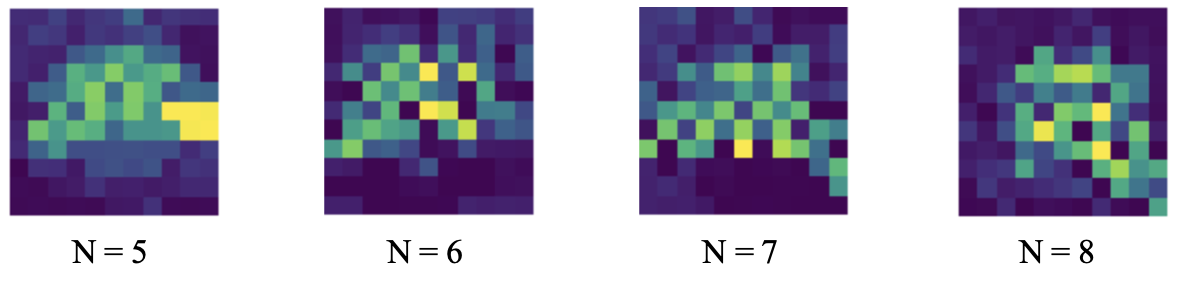}
  \caption{Representative synthetic rare-event image components generated for the four window sizes.}
  \label{fig:synthetic}
\end{figure}

Figures~\ref{fig:deepinsight} and~\ref{fig:synthetic} provide qualitative examples of the image modality. Because DeepInsight uses a fixed feature-to-pixel arrangement, differences across images reflect changes in market variables, while the generated examples illustrate the spatial patterns produced under rare-event conditioning.

\subsection{Runtime}

For $N=5$, all methods were measured on the same NVIDIA A100 GPU, excluding preprocessing and evaluation. TabSyn required 643.01 seconds for training and 5.47 seconds for sampling; TabDDPM required 463.38 seconds and 14.34 seconds, respectively. The multimodal framework required 697.87 seconds for training and 3.25 seconds for sampling. Its extra training cost reflects the three modality-specific encoders, while latent-space diffusion provides the fastest sampling of the compared methods.

\end{document}